\documentclass[conference]{IEEEtran}
\IEEEoverridecommandlockouts
\usepackage{cite}
\usepackage{amsmath,amssymb,amsfonts}
\usepackage{algorithmic}
\usepackage{graphicx}
\usepackage{textcomp}
\usepackage{xcolor}
\usepackage{multirow}
\usepackage{hyperref}
\usepackage{subcaption}
\usepackage{tikz}
\usetikzlibrary{shapes.geometric, arrows}
\usetikzlibrary{calc}

\def\BibTeX{{\rm B\kern-.05em{\sc i\kern-.025em b}\kern-.08em
    T\kern-.1667em\lower.7ex\hbox{E}\kern-.125emX}}
\begin{document}

\title{Data Augmentation for Graph Data: Recent Advancements\\}

\author{\IEEEauthorblockN{Maria Marrium}
\IEEEauthorblockA{\textit{Dept. of Computer Science} \\
\textit{Information Technology University}\\
Lahore, Pakistan \\
msds20011@itu.edu.pk}
\and
\IEEEauthorblockN{Arif Mahmood}
\IEEEauthorblockA{\textit{Dept. of Computer Science} \\
\textit{Information Technology University}\\
Lahore, Pakistan\\
arif.mahmood@itu.edu.pk}
}

\maketitle

\begin{abstract}
Graph Neural Network (GNNs) based methods have recently become a popular tool to deal with graph data because of their ability to incorporate structural information. The only hurdle in the performance of GNNs is the lack of labeled data. Data Augmentation techniques for images and text data can not be used for graph data because of the complex and non-euclidean structure of graph data. This gap has forced researchers to shift their focus towards the development of data augmentation techniques for graph data. Most of the proposed Graph Data Augmentation (GDA) techniques are task-specific. In this paper, we survey the existing GDA techniques based on different graph tasks. This survey not only provides a reference to the research community of GDA but also provides the necessary information to the researchers of other domains. 
\end{abstract}

\begin{IEEEkeywords}
 Graph data augmentation
\end{IEEEkeywords}

\section{Introduction}
\begin{figure*}
    \centering
    \includegraphics[width=1\linewidth]{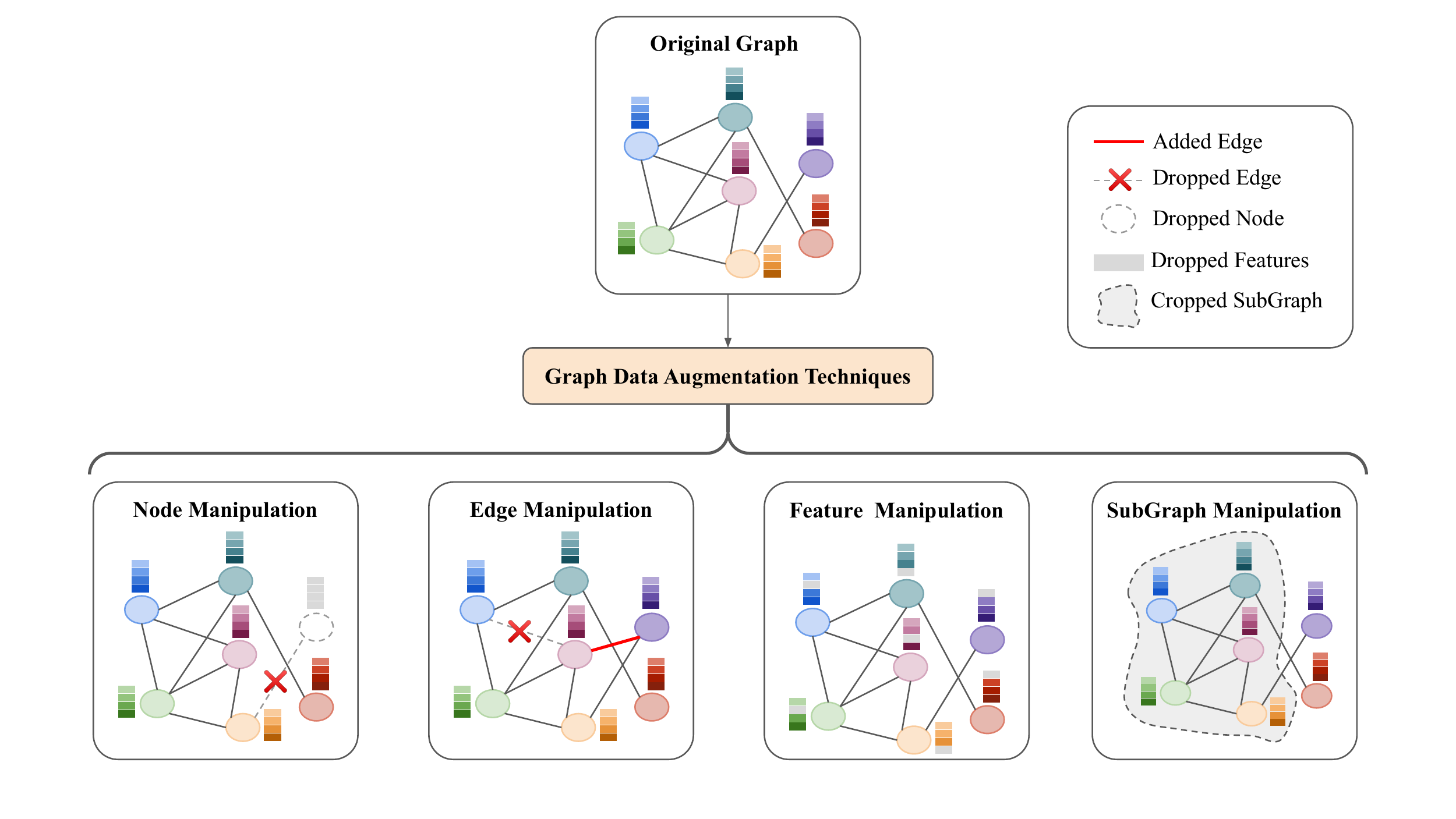}
    \caption{Graph Data Augmentation (GDA) Techniques for GNNs}
    \label{fig:exp}
\end{figure*}

\begin{figure*}
    \centering
    \includegraphics[width=1\linewidth]{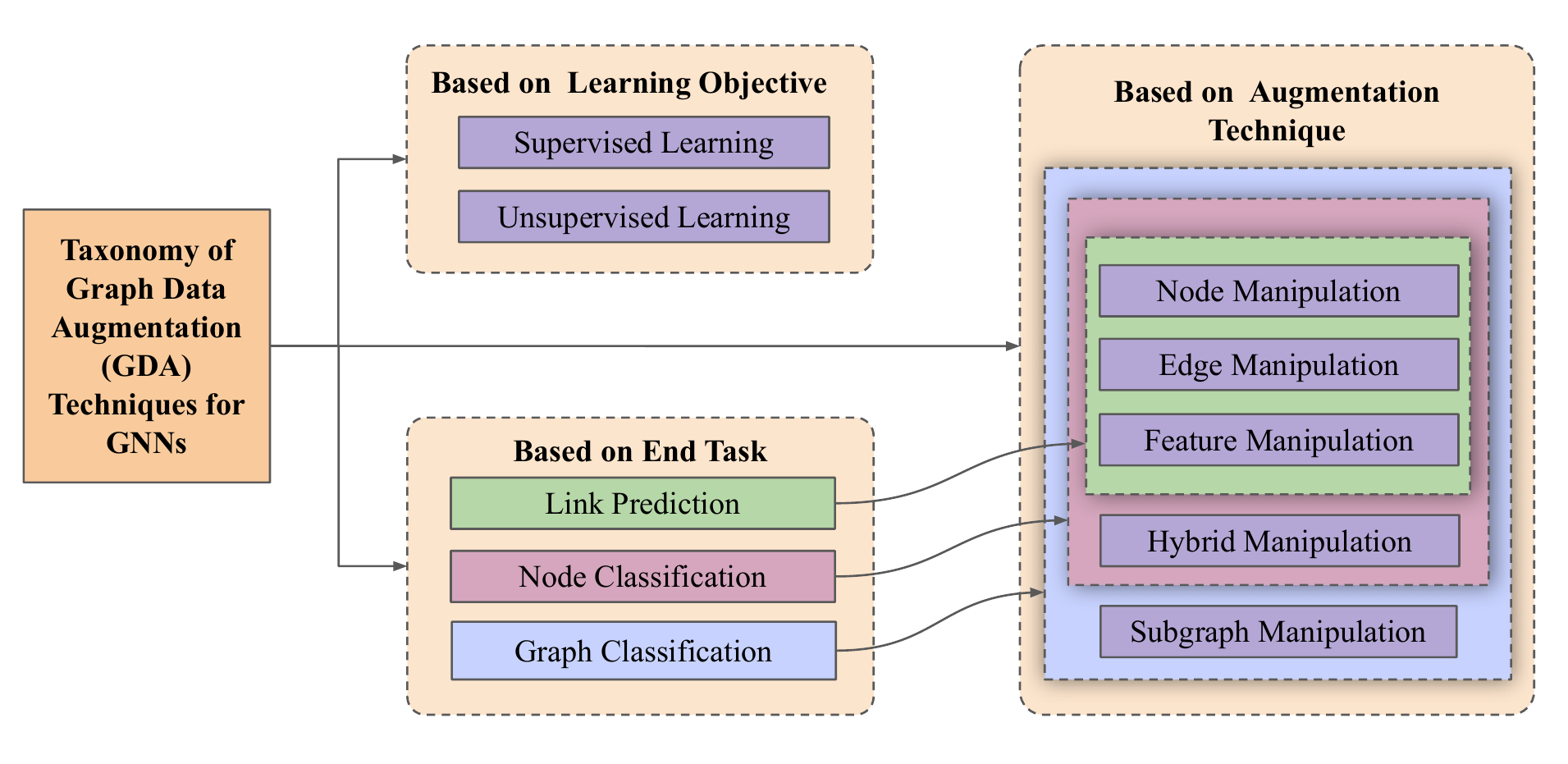}
    \caption{Taxonomy of Graph Data Augmentation (GDA) Methods for GNNs}
    \label{fig:GDA types}
\end{figure*}

Graph Neural Networks (GNNs) based methods have recently gained prominence because of their superior performance on graph data when compared to standard embedding-based approaches. 
The power of GNNs lies in their ability to model the structural information. GNNs are being widely used to perform tasks at three different levels including node, edge and graph. 
Node-level tasks such as node classification and missing data estimation aims to predict the label or missing data at a node in a graph. For example, predicting the class of a research paper in a citation network. 

Edge-level tasks or edge prediction's main objective is to predict whether there exists an edge between two nodes or not. For example, in recommendation system presence of an edge between two nodes will represent recommendation. 
Similarly, Graph-level tasks or graph classification goal is to predict the overall label of a graph. For example, given a chemical compound represented as a graph (nodes are atoms and edges represents the bond between them) classify it as toxic or non-toxic.
Since GNNs make data-driven inferences and like all other data-driven models their performance can be boosted by having more data. One can always collect and label more data to increase the size of the training dataset. However, gathering and labeling data is expensive. 
Data augmentations are the techniques used to increase the amount of labeled data by creating slightly modified copies or new synthetic data from the existing labeled data. 
These techniques are inexpensive compared to gathering and labeling new data in most cases. 
In the fields of Machine Learning (ML) and Natural Language Processing (NLP), data augmentation has been frequently used to expand the amount of training data.
Unfortunately, these techniques cannot be directly used for graph data owing to their complex and non-Euclidean structure. 
In the recent years many researchers have focused on developing data augmentation techniques for GNNs resulting in an increasing number of graph data augmentation methods. In the current work, we survey the existing data augmentation works for GNNs and provide taxonomies for them according to different three different criteria. These criteria are summarized in Figure \ref{fig:GDA types}. Next three paragraphs briefly explain these taxonomies. 

The graph data augmentation methods can be broadly divided into five classes based on augmentation techniques. These include  node manipulation i.e. creation or removal of nodes from graphs \cite{verma2019graphmix}, edge manipulation i.e. adding or dropping edges from a graph \cite{gao2021training, chen2020measuring}. The feature manipulation i.e. creating new features or modifying existing node/edge/graph features \cite{kong2020flag,liu2021local}. Sub-graph manipulation is for graph-level tasks only and refers to cropping/modifying/swapping sub-graphs within a graph \cite{park2021graph,sun2021mocl}. Figure \ref{fig:exp}, visually explains these techniques.  Hybrid manipulation is just a mixture of two or more aforementioned augmentation techniques \cite{sun2021automated,you2021graph}. 
Also, based on learning objective of GNNs GDA methods they can be classified into two classes including supervised learning \cite{tang2021data, spinelli2021fairdrop, zheng2020robust, zhao2020data} and unsupervised learning \cite{wang2020nodeaug, you2020graph,you2021graph,zhu2020deep}. 

Similarly, based on end task GDA methods can be divided into three classes: link prediction \cite{zhao2021counterfactual, kong2020flag, verma2019graphmix,wang2021adaptive}, node classification \cite{jin2020graph, wang2020nodeaug, sun2021automated} and graph classification \cite{wang2020graphcrop, kong2020flag,park2021graph}.
For link prediction and node classification task, a single graph is used for both training and inference. Augmentation methods for these two tasks perform data augmentation by manipulating nodes/edges/features or combination of these operations. For example, Wang \textit{et al.} \cite{wang2021adaptive} proposed data augmentation method for link prediction that generates multiple augmented graphs by perturbing time and edges and then a multi-level module uses these graphs to make predictions. Rong \textit{et al.} \cite{rong2019dropedge} proposed data augmentation algorithm for node classification which randomly drops edges during training. 
For graph-level tasks we have multiple graphs and usually different set of graphs are used at training and inference stage. So, data augmentation techniques for graph-level tasks  generate new graphs from the limited number of available graph data by manipulating a subset of nodes/edges/features/sub-graph or any combination of these operations to improve generalization ability. 
For example, Wang \textit{et al.} \cite{wang2020graphcrop} first randomly selected an initial node and then generated new graphs by retaining the strongly connected sub-graph with the initial node.

\begin{figure*}
    \centering
    \begin{tikzpicture}[node distance=2cm]
    \tikzset{every node/.style= {thick, draw=black, align=center, minimum height=18pt, text width= 100pt}}
    
    \node[fill=blue!20](n1) {For Node Classification};
    \node[fill=green!40, right=20pt,below=10pt] (n2) at (n1.south) {DropEdge (2019) \cite{rong2019dropedge}};
    \node[fill=green!40, text width= 120pt, right=20pt,below=10pt] (n3) at (n2.south) {NeuralSparse (2020) \cite{zheng2020robust}};
    \node[fill=green!40,right=20pt,below=10pt] (n4) at (n3.south) {TADropEdge (2021)\cite{gao2021training}};
    \node[fill=green!40,below=65pt] (n5) at (n2.south) {BGCN (2019)\cite{zhang2019bayesian}};
    \node[fill=green!40,below=90pt] (n6) at (n2.south) {GDC (2019)\cite{klicpera2019diffusion}};
    \node[fill=pink!50,below=115pt] (n7) at (n2.south) {DGI (2019)\cite{velicokovic2018deep}};
    \node[fill=green!40,below=140pt] (n8) at (n2.south) {Pro-GNN(2020)\cite{jin2020graph}};
    \node[fill=green!40,below=165pt] (n9) at (n2.south) {AdaEdge (2020) \cite{chen2020measuring}};
    \node[fill=green!40,right=20pt,below=10pt] (n10) at (n9.south) {GAug (2021) \cite{zhao2021data}};
    \node[fill=red!40, right=20pt,below=10pt] (n11) at (n10.south) {AutoGRL (2021) \cite{sun2021automated}};
    \node[fill=red!40,below=250pt] (n12) at (n2.south) {NodeAug (2020)\cite{wang2020nodeaug}};
    \node[fill=yellow!40,below=275pt] (n13) at (n2.south) {GraphMix (2020)\cite{verma2019graphmix}};
    
    \node[fill=yellow!40,text width= 110pt, right=20pt,below=10pt] (n14) at (n13.south) {Graph Mixup (2021) \cite{wang2021mixup}};
    \node[fill=red!40, right=20pt,below=10pt] (n15) at (n14.south) {NodeAug-I (2021)\cite{xue2021node}};
    
    \node[fill=pink!50,below=360pt] (n16) at (n2.south) {FLAG (2020) \cite{kong2020flag}};
    \node[fill=red!40,below=385pt] (n17) at (n2.south) {GRAND (2020) \cite{feng2020graph}};
    \node[fill=red!40,below=410pt] (n18) at (n2.south) {GRACE (2020) \cite{zhu2020deep}};
    \node[fill=green!40,below=435pt] (n19) at (n2.south) {FairDrop (2021) \cite{spinelli2021fairdrop}};
    \node[fill=pink!50,below=460pt] (n20) at (n2.south) {LA-GNN (2021) \cite{liu2021local}};
    \node[fill=pink!40,below=485pt, text width= 110pt] (n21) at (n2.south) {Tang et al. (2021) \cite{tang2021data}};
    \node[fill=green!40,below=510pt] (n22) at (n2.south) {MV-GCN (2021) \cite{yuan2021semi}};

    \node[fill=blue!20,right=100pt] (g1) at (n1.east) {For Graph Classification};
    \node[fill=orange!40,right=20pt, below=10pt] (g2) at (g1.south) {GraphCrop (2020)\cite{wang2020graphcrop}};
    \node[fill=pink!50,right=20pt, below=35pt] (g3) at (g1.south) {FLAG (2020)\cite{kong2020flag}};
    \node[fill=green!40,right=20pt,below=60pt] (g4) at (g1.south) {M-Evolve (2020)\cite{zhou2020data}};
   \node[fill=red!40,right=20pt,below=85pt] (g5) at (g1.south) {GrapCL (2020)\cite{you2020graph}};
    
    \node[fill=orange!40,right=20pt,below=110pt] (g6) at (g1.south) {MoCL (2021)\cite{sun2021mocl}};
    \node[fill=red!40,right=20pt,below=135pt] (g7) at (g1.south) {JOAO (2021)\cite{you2021graph}};
    \node[fill=orange!40,fill=yellow!40,text width= 105pt,below=10pt] (g8) at (g7.south) {Graph Mixup (2021) \cite{wang2021mixup}};
    \node[fill=orange!40, right=20pt, below=10pt] (g9) at (g8.south) {ifMixup(2021) \cite{guo2021ifmixup}};
    
    \node[fill=orange!40,text width= 122pt, right=20pt, below=10pt] (g10) at (g9.south) {Graph Transplant (2021) \cite{park2021graph}};
    \node[fill=orange!40,right=5pt, below=10pt] (g11) at (g10.south) {G-Mixup (2022) \cite{han2022g}};
    \node[fill=orange!40,right=20pt, below=275pt] (g12) at (g1.south) {GAMS (2022)\cite{bicciato2022gams}};

    \node[fill=blue!20,right=270pt] (e1) at (n1.east) {For Link Prediction};
    \node[fill=yellow!40,right=20pt, below=10pt] (e2) at (e1.south) {GraphMix (2020)\cite{verma2019graphmix}};
    \node[fill=pink!50,below=10pt] (e3) at (e2.south) {FLAG (2020)\cite{kong2020flag}};
    \node[fill=green!40,below=10pt] (e4) at (e3.south) {MeTA (2021)\cite{wang2021adaptive}};
    \node[fill=green!40,below=10pt] (e5) at (e4.south) {CFLP (2021)\cite{zhao2021counterfactual}};
    
    \node[fill=blue!60, text width= 200pt, above=30pt, right=100pt](top) at (n1.north){\textbf{Data Augmentation Techniques for GNNs}};
    
    \draw[thick] (10,-17.7) rectangle (15.5,-13);
    \draw[thick, fill=green!40] (10.5,-13.5) rectangle (11.5,-14);
    \draw[thick, fill=yellow!40] (10.5,-14.3) rectangle (11.5,-14.8);
    \draw[thick, fill=red!40] (10.5,-15.1) rectangle (11.5,-15.6);
    \draw[thick, fill=pink!50] (10.5,-15.9) rectangle (11.5,-16.4);
    \draw[thick, fill=orange!40] (10.5,-16.7) rectangle (11.5,-17.2);
    
    \node[draw=none, left= 15pt, below=258pt,align=left] (l1) at (e5.south) {Edge Augmentation};
    
    \node[draw=none, below=4pt, align=left] (l2) at (l1.south) {Node Augmentation};
    \node[draw=none, below=4pt, align=left] (l3) at (l2.south) {Hybrid Augmentation};
    \node[draw=none, below=4pt, align=left] (l4) at (l3.south) {Feature Augmentation};
    \node[draw=none, right=35pt, below=4pt, align=left, text width= 170pt] (l5) at (l4.south) {Sub-graph Augmentation};

    \coordinate (atop) at ($(top.south) + (0,-10pt)$);
    
    \draw[thick] (top.south) -- (top.south|-g1.north)
    (n1.north) |- (atop) -| (e1.north)
    (n1.south west) |- (n2.west)
    (n1.south west) |- (n5.west)
    (n1.south west) |- (n6.west)
    (n1.south west) |- (n7.west)
    (n1.south west) |- (n8.west)
    (n1.south west) |- (n9.west)
    (n1.south west) |- (n12.west)
    (n1.south west) |- (n13.west)
    (n1.south west) |- (n16.west)
    (n1.south west) |- (n17.west)
    (n1.south west) |- (n18.west)
    (n1.south west) |- (n19.west)
    (n1.south west) |- (n20.west)
    (n1.south west) |- (n21.west)
    (n1.south west) |- (n22.west)
    
    (n2.south west) |- (n3.west)
    (n3.south west) |- (n4.west)
    (n9.south west) |- (n10.west)
    (n10.south west) |- (n11.west)
    (n13.south west) |- (n14.west)
    (n14.south west) |- (n15.west)

    (g1.south west) |- (g2.west)
    (g1.south west) |- (g3.west)
    (g1.south west) |- (g4.west)
    (g1.south west) |- (g5.west)
    (g1.south west) |- (g6.west)
    (g1.south west) |- (g7.west)
    (g1.south west) |- (g8.west)
    (g1.south west) |- (g12.west)
    (g8.south west) |- (g9.west)
    (g9.south west) |- (g10.west)
    (g10.south west) |- (g11.west)
    
    (e1.south west) |- (e2.west)
    (e1.south west) |- (e3.west)
    (e1.south west) |- (e4.west)
    (e1.south west) |- (e5.west);

    \end{tikzpicture}
    \caption{Data Augmentation Techniques for GNNs.}
    
    \label{fig:GDA Tech}
\end{figure*}
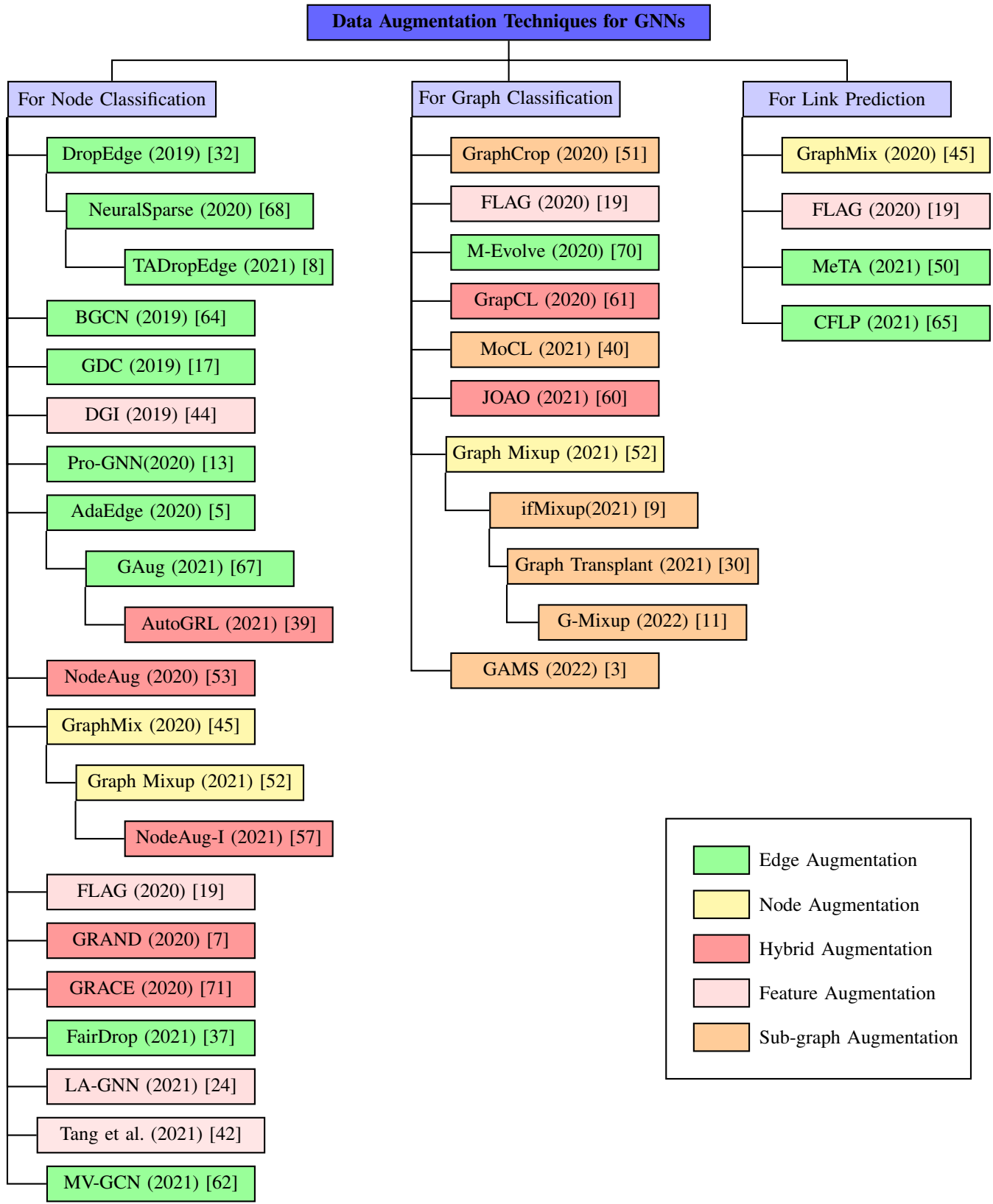

Overall the main contributions of this work are as follows:
\begin{itemize}
    \item For the new readers to quickly understand the similarities and differences between different works, we summarizes the existing work on the basis of three different criteria including the end task to be accomplished, augmentation technique and learning objective. Figure \ref{fig:GDA Tech} summarizes the existing works in chronological order. Table \ref{tab:paper_div} \& \ref{tab:paper_div_aug_tech} provides the list of papers divided by these taxonomies.
    
    \item Due to  dataset variations and large number of evaluation metrics it becomes hard to compare the results of two different augmentation techniques. Our work facilitates to answer the question: which data augmentation technique works best for a specific task or on a specific dataset? The only way to do so is by building benchmarks similar to the other areas \cite{deng2009imagenet, wang2018glue}. 
    
    \item We summarizes the frequently used datasets and evaluation metrics for all papers. We hope this work would help community to build a good benchmark for this domain.
    
    \item To help researchers of this domain or individuals in general we also list the open-source implementations of the existing works in table \ref{tab:links}.
\end{itemize}

The rest of the paper is organized as follows. Section \ref{sec:background}, provides some necessary background information about graphs and GNNs. Section \ref{sec:node-aug}, summarizes the node-level GDA methods. Section \ref{sec:garph-aug}, lists the graph-level GDA methods. Section \ref{sec:edge-aug}, briefly describes the edge-level GDA methods. Section \ref{sec:data and eval} provides the details of datasets and evaluation metrics being commonly used. For ease we divide them by task-level. Section \ref{sec:perform comp} compares the performance of existing GDA methods. The last section provides future research directions and concludes this survey. In the appendix \ref{appendix}, we provide the the links of open-source implementation of the existing works. 
\begin{table*}[]
    \centering
    \caption{Graph Data Augmentation Works Categorized on the basis of Learning Objective and End Task}
    
    \begin{tabular}{l|ll}
    \hline
         \textbf{Learning objective} & \textbf{End Task} & \textbf{Paper} \\
         \hline
         \multirow{3}{4em}{Supervised Learning} 
         & Link Prediction & \cite{zhao2021counterfactual,wang2021adaptive,kong2020flag}\\[1ex]
         \cline{2-3}
         & Node Classification & \cite{rong2019dropedge,wang2021mixup,kong2020flag,klicpera2019diffusion, tang2021data,spinelli2021fairdrop,verma2019graphmix,zheng2020robust,gao2021training,spinelli2021fairdrop,yuan2021semi,zhao2020data,chen2020measuring,jin2020graph,wang2021mixup,sun2021automated, feng2020graph}\\[1ex]
         \cline{2-3} 
         
         & Graph Classification & \cite{wang2020graphcrop,zhou2020data,sun2021mocl,wang2021mixup,kong2020flag,guo2021ifmixup,park2021graph}\\[1ex]
         \hline
         \multirow{3}{4em}{Unsupervised Learning} & Link Prediction & -\\
         \cline{2-3}
         & Node Classification & \cite{wang2020nodeaug, velicokovic2018deep, zhu2020deep}\\[1ex]
         \cline{2-3}
         & Graph Classification & \cite{you2020graph,you2021graph}\\[1ex]
         \hline

    \end{tabular}
    \label{tab:paper_div}
\end{table*}

\begin{table*}[]
    \centering
    \caption{Graph Data Augmentation Works Categorized on the basis of End Task and Augmentation Technique}
    \begin{tabular}{p{2cm}|p{3cm}l}
    \hline
         \textbf{End Task} & \textbf{Augmentation Technique} & \textbf{Paper}
         \\
         \hline
         \multirow{5}{1em}{Link Prediction} & Edge Manipulation & \cite{wang2021adaptive, zhao2021counterfactual}\\
         \cline{2-3}
        & Node Manipulation &  \cite{verma2019graphmix} \\
        \cline{2-3}
         & Feature Manipulation & \cite{kong2020flag}\\
         \cline{2-3}
         & Sub-Graph Manipulation &  - \\
         \cline{2-3}
         & Hybrid Manipulation & - \\
         \hline

         \multirow{5}{4em}{Node Classification} & Edge Manipulation & \cite{rong2019dropedge, jin2020graph, chen2020measuring, zheng2020robust, gao2021training,spinelli2021fairdrop,klicpera2019diffusion,zhao2021data, yuan2021semi, zhang2019bayesian}\\
         \cline{2-3}
         & Node Manipulation & \cite{verma2019graphmix, wang2021mixup}\\
         \cline{2-3}
         & Feature Manipulation & \cite{kong2020flag, tang2021data,liu2021local, velicokovic2018deep}\\
         \cline{2-3}
         & Sub-Graph Manipulation &  - \\
         \cline{2-3}
         & Hybrid Manipulation & \cite{sun2021automated, wang2020nodeaug, feng2020graph, xue2021node, zhu2020deep}\\
         \hline
         \multirow{5}{4em}{Graph Classification} & Edge Manipulation & \cite{zhou2020data}\\
         \cline{2-3}
         & Node Manipulation & \cite{wang2021mixup} \\
         \cline{2-3}
         & Feature Manipulation & \cite{kong2020flag}\\
         \cline{2-3}
         & Sub-Graph Manipulation & \cite{wang2020graphcrop, sun2021mocl, guo2021ifmixup, park2021graph, bicciato2022gams, han2022g}\\
         \cline{2-3}
         & Hybrid Manipulation & \cite{you2020graph, you2021graph}\\
         \hline
    \end{tabular}
    
    \label{tab:paper_div_aug_tech}
\end{table*}
\section{Background}\label{sec:background}
In this section, we discuss notation for graph data and different types of graph data with examples.

\subsection{Graphs}
We denote a set of graphs as $G=\{G^p\}_{p=1}^ n$, where $|G|=N$. Each graph $G^p = (V^p,A^p)$ where $V^p$ is the set of $m$ nodes $\{v^p_1, v^p_2, ... v^p_m\}$. $W^p$ is $m \times m$ matrix containing the edge weights $W_{i,j} \neq 0 $ if there is an edge between $v^p_i$ and $v^p_j$, and 0 otherwise. $X_p$ is a Nxd feature matrix, where $m^{th}$ row of $X_i$ denotes node feature of $v_m$. 

\subsection{Types of Graphs}
\textbf{Directed Graphs Vs. Undirected Graphs.} A directed graph has direction information associated with each edge. Consider, $e^p_{i,j} = (v^p_i, v^p_j)$ an edge between node $v^p_i$ and $v^p_j$ and  $e^p_{j,i} = (v^p_j, v^p_i)$ an edge between node $v^p_j$ and $v^p_i$. For directed graphs $e^p_{i,j} \neq e^p_{j,i}$. An example of a directed graph is Twitter, where each node is a user and an edge represents the follow-ship information. In undirected graphs, any two nodes  share the same edge: $e^p_{i,j} = e^p_{j,i}$. For example, Facebook friendships is an undirected graph where nodes are users and the edge represents friendship. 

\textbf{Dynamic Graphs Vs. Static Graphs.} A dynamic graph is the one whose nodes, edges, node features or edge features changes over time. Social networks are good examples of dynamic networks. A static graph has fixed set of node, edge, node features and edge features. A common example of static graph is molecular structure of drugs.
\subsection{Graph Neural Networks}
Graph Neural Networks (GNNs) are a class of deep learning methods that are designed to perform node, edge and graph level inference on graph data. GNNs encode node features into low-dimensional to learn representation vectors $h_{G}$ and $h_{v}$ for  entire graph and for every node $v \in V$ respectively.  Here we present the formal expression of Graph Convolution Network (GCN) \cite{kipf2016semi}. The k-th layer of GCN is described as following:
\begin{equation}
\begin{split}
    msg_{v}^{(k)} = AGG^{(k)}(\{h_j^{(k-1)}:j \in N(v)\}) \\
    h_{v}^{(k)} = COMBINE^{(k)} (h_v^{(k-1)},msg_{v}^{(k)}) 
\end{split}
\end{equation}
where $h_{v}^{k} \in \mathbb{R}^{n\times d_{k}}$ is the node representation at k-th layer. $N(v)$ represents the neighbors of node $v$. $AGG(.)$ is the aggregation function to collect node information via message passing. COMBINE(.) combines the neighbors representation at k-th layer and node representation at (k-1)-th layer. Graph representation $h_{G}$ is obtained by using permutation-invariant READOUT(.) function. This function pools node features from the the final iteration K as follows:
\begin{equation}
    h_{G} = READOUT(\{ h_{v}^{(K)} | v \in V\})
\end{equation}

\begin{table*}[]
    \centering
    \caption{Summary of Datasets for Node Classification and Link Prediction}
    \begin{tabular}{l|p{2cm}|lccccp{3cm}}
    \hline
         \textbf{Type} & \textbf{Task-Level} & \textbf{Datasets}& \textbf{Source}  & \textbf{\#Nodes} & \textbf{\#Edges} & \textbf{\#Classes} & \textbf{Paper}  \\
         \hline
         \multirow{5}{4em}{Citation} & Node/Edge & Cora & \cite{sen:aimag08} &  2,708 & 5,429 & 7 & \cite{chen2020measuring, sun2021automated,zhang2019bayesian,zhao2021counterfactual,rong2019dropedge,kong2020flag,zhao2020data,klicpera2019diffusion,wang2021mixup,verma2019graphmix, liu2021local, yuan2021semi, zheng2020robust, xue2021node, wang2020nodeaug, jin2020graph, gao2021training, feng2020graph,velicokovic2018deep}\\
         \cline{2-8}
         &Node/Edge & Citeseer &  \cite{sen:aimag08}&  3,327 & 4,732 & 6& \cite{chen2020measuring,zhang2019bayesian,zhao2021counterfactual,tang2021data,rong2019dropedge,spinelli2021fairdrop,zhao2020data,klicpera2019diffusion,wang2021mixup,verma2019graphmix, liu2021local,yuan2021semi,zheng2020robust,xue2021node, wang2020nodeaug, jin2020graph, gao2021training, feng2020graph, velicokovic2018deep}\\ 
        \cline{2-8}
        &Node/Edge & Pubmed & \cite{sen:aimag08} &  19,717 & 44,338 & 3 & \cite{chen2020measuring,zhang2019bayesian,zhao2021counterfactual,rong2019dropedge,spinelli2021fairdrop,klicpera2019diffusion, wang2021mixup,verma2019graphmix, liu2021local, yuan2021semi, xue2021node, wang2020nodeaug, jin2020graph, gao2021training, feng2020graph, velicokovic2018deep}\\
        \cline{2-8}
        & Node/Edge& Cora-ML & \cite{mccallum2000automating} & 2,810 & 7981 & 7 & \cite{spinelli2021fairdrop}\\
        \cline{2-8}
        &Node & OGB-arxiv & \cite{NEURIPS2020_fb60d411} & 169,343 & 1,166,243 & & \cite{kong2020flag}\\
         \hline
        \multirow{4}{4em}{Coauthor} & Node & CS & & 18,333& 81,894& 15 & \cite{chen2020measuring, klicpera2019diffusion,verma2019graphmix, yuan2021semi, wang2020nodeaug}\\
        \cline{2-8}
        & Node & Physics& &  34,493 & 247,962 & 5 & \cite{chen2020measuring,verma2019graphmix, wang2020nodeaug}\\
        \cline{2-8}
        & Node & ACM & \cite{10.1145/3308558.3313562}& 3,025 & 13128 & 3 & \cite{tang2021data}\\
        \cline{2-8}
        & Node & DBLP &\cite{pmlr-v119-buyl20a} & 3980 & 6965& 5 & \cite{spinelli2021fairdrop}\\
        \cline{2-8}
        \hline
        Traffic & Node & USA-Airports& & 1,190 & 13,599 & 4 &\cite{sun2021automated,zhao2020data}\\
        \hline
        Web & Node & WikiCS & & 10,711 & 216,123&&\cite{sun2021automated}\\
        \hline
        \multirow{4}{4em}{Co-Purchasing} & Node & Amazon Computers & & 13,381 & 245,778 & 10 & \cite{chen2020measuring,klicpera2019diffusion, yuan2021semi, xue2021node}\\
        \cline{2-8}
        & Node & Amazon Photos& &  7,487 & 119,043 & 8 & \cite{chen2020measuring,sun2021automated, klicpera2019diffusion, yuan2021semi, xue2021node}\\
        \cline{2-8}
        & Node & OGB-Products & \cite{NEURIPS2020_fb60d411}& 2,449,029 & 61,859,140& & \cite{kong2020flag}\\
        \cline{2-8}
        & Node & Amazon &\cite{mcauley2012image} & 1,598,960 & 132,169,734 & 107 & \cite{wang2021mixup}\\
        \hline
        \multirow{6}{4em}{Social} & Node/Edge & Facebook & \cite{NIPS2012_7a614fd0} & 4,039& 88,234& 2 &  \cite{zhao2021counterfactual,spinelli2021fairdrop}\\
        \cline{2-8}
        & Node & BlogCatalog &\cite{10.1145/3289600.3291015} & 5,196& 1,71,743 & 6 & \cite{tang2021data,zhao2020data}\\
        \cline{2-8}
        & Node & Flickr &\cite{10.1145/3289600.3291015} & 7,575& 2,39,738 & 9 & \cite{tang2021data,zhao2020data,wang2021mixup}\\
        \cline{2-8}
        & Node/Edge & Reddit &\cite{baumgartner2020pushshift} & 232,965 & 11,606,919 & 41 & \cite{rong2019dropedge, wang2021adaptive, zheng2020robust, velicokovic2018deep}\\
        \cline{2-8}
        & Node & Yelp & \cite{mcauley2012image} & 716,847 & 6,977,410 & 100 & \cite{wang2021mixup}\\
        \cline{2-8}
       & Node & Polblogs &\cite{adamic2005political} & 1,222 & 16,714 & 2 & \cite{jin2020graph}\\
        
        \hline
        Drugs Interaction & Edge & OGB-DDI & \cite{NEURIPS2020_fb60d411} & 4,267& 1,334,889 & - & \cite{zhao2021counterfactual,kong2020flag,han2022g}\\
        \hline
        
        \multirow{2}{4em}{Proteins Interaction} & Node & OGB-Proteins & \cite{NEURIPS2020_fb60d411} & 132,534 & 39,561,252& & \cite{kong2020flag}\\
        \cline{2-8}
        & Node & PPI & \cite{hamilton2017inductive} & 10,076 & 157,213 & 121& \cite{zhao2020data, zheng2020robust, velicokovic2018deep}\\
    
        \hline
        Collaboration & Edge & OGB-Collab & \cite{NEURIPS2020_fb60d411}& 235,868 & 1,285,465 &-& \cite{kong2020flag}\\
        \hline

        \multirow{9}{4em}{Other} & Node & OGB-mag & \cite{NEURIPS2020_fb60d411} & 1,939,743 & 21,111,007 & & \cite{kong2020flag}\\
        \cline{2-8}
        & Node & UAI2010 & \cite{10.1007/978-3-319-93040-4_18} & 3,067 & 28,311 & 19 & \cite{tang2021data}\\
        \cline{2-8}
        & Edge & Bitcoin Alpha & \cite{kumar2016edge} & 3,783 & 24,186 & 2 & \cite{verma2019graphmix}\\
        \cline{2-8}
        & Edge & Bitcoin OTC &\cite{kumar2016edge} & 5,881 & 35,592 & 2 & \cite{verma2019graphmix}\\
        \cline{2-8}
        & Node & Squirrel & \cite{rozemberczki2021multi} & 5201 & 217,073 & 5 & \cite{liu2021local} \\ 
        \cline{2-8}
        & Node & Actor &\cite{tang2009social} & 7600 & 33,544 & 5&\cite{liu2021local} \\ 
        \cline{2-8}
        & Node & Chameleon & \cite{rozemberczki2021multi}& 2277 & 36101 & 5 & \cite{liu2021local} \\ 
         \cline{2-8}
        & Edge & MOOC &\cite{kumar2019predicting} & 7,144 & 411,749 & &  \cite{wang2021adaptive}\\
         \cline{2-8}
        & Edge & Wikipedia & \cite{kumar2019predicting} & 9,277 & 157,474 & - &\cite{wang2021adaptive}\\
        
        \hline
    \end{tabular}
    \label{tab:node/edge data}
\end{table*}

\begin{table*}[]
    \centering
    \caption{Summary of Datasets for Graph Classification}
    \begin{tabular}{l|lcccccp{3cm}}
    \hline
         \textbf{Type} & \textbf{Dataset} & \textbf{Source} & \textbf{\#Graphs}& \textbf{Aug. \#Nodes} & \textbf{Avg. \#Edges} & \textbf{\#Classes} & \textbf{Paper}\\
        \hline
         \multirow{3}{4em}{Social} & Reddit-Binary & \cite{baumgartner2020pushshift}& 2000 & 429.63& 497.75& 2 & \cite{bicciato2022gams,han2022g,you2020graph,wang2020graphcrop, you2021graph}\\
        \cline{2-8}
         & Reddit-5K & \cite{baumgartner2020pushshift}& 4999& 508.52 & 594.87 & 5 & \cite{han2022g,wang2021adaptive,wang2020graphcrop, you2021graph}\\
          \cline{2-8}
         & Reddit-12K & \cite{baumgartner2020pushshift} & 11929& 391.41 & 456.89 & 11 & \cite{han2022g}\\
        \hline
        
        \multirow{12}{4em}{Chemical Compounds}& HIV & \cite{NEURIPS2020_fb60d411} & 41,127& 25.5 & 27.5 & 2& \cite{kong2020flag,han2022g, you2021graph}\\
         \cline{2-8}
         & bace & \cite{subramanian2016computational}& 1,513&34.1 & 36.9 & 2 & \cite{han2022g, you2021graph, sun2021mocl}\\
         \cline{2-8}
         & pcba & \cite{NEURIPS2020_fb60d411}&437,929 & 26.0 & 28.1 & & \cite{kong2020flag}\\
         \cline{2-8}
        
        & tox21 & \cite{novick2013sweetlead}& 7,831 & 18.6 & 19.3 & 2 & \cite{you2021graph, sun2021mocl}\\
        \cline{2-8}
        & bbbp & \cite{NEURIPS2020_fb60d411}& 2,039&24.1 & 26.0 & 2 & \cite{han2022g, you2021graph, sun2021mocl}\\
         \cline{2-8}
        & sider & \cite{kuhn2016sider}& 1,427 & 33.6 & 35.4 & 2 &  \cite{you2021graph,sun2021mocl}\\
        \cline{2-8}
        & toxcast & \cite{NEURIPS2020_fb60d411}& 8,576 & 18.8 & 19.3 & 2 & \cite{you2021graph, sun2021mocl}\\
        \cline{2-8}
        & MUTAG &\cite{kazius2005derivation} & 188 & 17.93 & 19.79 & 2 & \cite{bicciato2022gams, park2021graph,guo2021ifmixup, you2021graph, zhou2020data, sun2021mocl}\\
        \cline{2-8}
        & clintox&\cite{NEURIPS2020_fb60d411}& 1,477 & 26.2 &  27.9 & 2 & \cite{you2021graph,sun2021mocl}\\
        \cline{2-8}
        & muv & \cite{NEURIPS2020_fb60d411}& 93,087 & 24.2 &  26.3 & 2 & \cite{you2021graph}\\
        \cline{2-8}
         
        & ENZYMES &\cite{schomburg2004brenda} & 600 & 32.63 & 62.14 & 6 & \cite{bicciato2022gams, park2021graph,wang2020graphcrop, guo2021ifmixup, zhou2020data}\\
        \cline{2-8}
        & PTC-MR & & 334 & 14.3 & 29.4 & 2& \cite{guo2021ifmixup, zhou2020data}\\
        \hline

        \multirow{2}{4em}{Co-featuring} &  IMDB-B & \cite{baumgartner2020pushshift} &1000 & 19.77 & 96.53 & 2 & \cite{han2022g,wang2020graphcrop,guo2021ifmixup, you2021graph}\\
         \cline{2-8}
        & IMDB-M & \cite{baumgartner2020pushshift}&1500 & 13.00 & 65.94 & 3 & \cite{han2022g, bicciato2022gams,wang2021mixup, wang2020graphcrop,guo2021ifmixup}\\
        \hline
        \multirow{2}{4em}{Proteins Interaction} & OGB-ppa & \cite{NEURIPS2020_fb60d411} &158,100 & 243.4& 2,266.1 & & \cite{kong2020flag, park2021graph, you2021graph}\\
         \cline{2-8}
        & Proteins & \cite{KKMMN2016} & 1113 & 39.06 & 72.82 & 2 & \cite{bicciato2022gams,wang2021mixup,you2020graph, wang2020graphcrop, guo2021ifmixup, you2021graph, han2022g}\\
        
        \hline
        Collaboration & Collab & \cite{yanardag2015deep} & 5000 & 74.49 & 2457.78 & 3 & \cite{bicciato2022gams, park2021graph,wang2021mixup,you2020graph,you2021graph}\\
        \hline
        
       \multirow{3}{4em}{Anti-Cancer Screen} & NCI-H23 & \cite{yan2008mining}&   & - & - & 2 & \cite{park2021graph}\\
        \cline{2-8}
        & MOLT-4 & \cite{yan2008mining}&  & - & - & 2 & \cite{park2021graph}\\
         \cline{2-8}
        & P388 & \cite{yan2008mining}& & - & - & 2 & \cite{park2021graph}\\
        
        \hline
        \multirow{3}{4em}{Brain} & KKI & \cite{KKMMN2016}& 83 & 26.96 & 48.42 &2 & \cite{zhao2020data}\\
        \cline{2-8}
        & Peking-1 &\cite{KKMMN2016} & 85 & 39.31 & 77.35 & 2 & \cite{zhao2020data}\\
        \cline{2-8}
        & OHSU &\cite{KKMMN2016} & 79 & 82.01 &  199.66 & 2 & \cite{zhao2020data}\\
        
        \hline
        \multirow{6}{4em}{Other} & OGB-code & \cite{NEURIPS2020_fb60d411} & 452,741&125.2& 124.2& & \cite{kong2020flag, you2021graph}\\
        \cline{2-8}
        & D\&D &\cite{KKMMN2016} & 1178 & 284.32 & 715.66 & 2 & \cite{bicciato2022gams,wang2021mixup,wang2020graphcrop, you2021graph}\\
         \cline{2-8}
        & MSRC-21 &\cite{KKMMN2016} & 563 & 77.52 & 198.32 & 20 & \cite{bicciato2022gams}\\
         \cline{2-8}
        & NCI1 &\cite{wale2008comparison} & 4110 &  29.87 & 32.30& 2 & \cite{bicciato2022gams, park2021graph, wang2021mixup,you2020graph,wang2020graphcrop,guo2021ifmixup, you2021graph}\\
         \cline{2-8}
        & NCI109 &\cite{KKMMN2016} & 4,127 & 29.68 & 32.13 & & \cite{wang2020graphcrop,guo2021ifmixup}\\
         \cline{2-8}
        & COIL-DEL & \cite{riesen2008iam} &3900 & 21.5 & 54.2 & 100 & \cite{park2021graph}\\
        \hline
    \end{tabular}
    
    \label{tab:graph-level data}
\end{table*}

\section{Graph Data Augmentation for Node Classification}\label{sec:node-aug}
Given a network $G = (V,A,X)$ and a subset of labelled nodes $V_L \subseteq V$ with class labels from $Y = \{1,2,...,K\}$, where K is the number of categories. The goal of the node classification algorithm is to learn a function $C\;:V\rightarrow Y$ that predicts the node label for each node $v \in V$. The function C is usually learned by minimizing the cross-entropy loss between the predicted label and the original label of labeled nodes $V_L$. Node-level tasks are the most popular ones.

\subsection{Data Augmentation Methods}
Present data augmentation techniques for node classification can be broadly divided into four classes: 1) Edge Manipulation and 2) Node Manipulation 3) Feature  Manipulation 4) Hybrid Manipulation.

\subsubsection{Edge Manipulation}
Edge manipulation-based data augmentation techniques aim to manipulate the existing graph topology by adding some perturbation to the adjacency matrix of the graph. 
Rong \textit{et al.} proposed DropEdge \cite{rong2019dropedge}  that randomly drops a fixed percentage of edges at each training epoch. Dropping edges slightly modify the original graph and thus GNN sees a different graph at each epoch. DropEdge significantly improves the GNN model generalization capability and also helps in reducing the over-smoothing problem of GNN (especially in deeper GNN). One of the major concerns about DropEdge technique is that it may remove task-relevant edges and decrease the overall performance of GNN model. 
To cater to this issue, Zheng \textit{et al.}  proposed NeuralSparse \cite{zheng2020robust} that uses MLP based model trained jointly with GNN on node classification loss and learns to drop only task-irrelevant edges. 
TADropEdge (Topology Adaptive Drop edge)\cite{gao2021training} proposed by Gao \textit{et al.} take into consideration the graph structural information when randomly dropping edges.  In this way, the proposed technique makes sure that the augmented graph preserves the topological structure of the original graph. 
Similarly, Spinelli \textit{et al.} \cite{spinelli2021fairdrop} proposed FairDrop that biasedly drop edges to reduce the negative effects of homophily sensitive attributes.

Another work by Klicpera \textit{et al.} proposed Graph Diffusion Convolution (GDC) \cite{klicpera2019diffusion}, that utilizes Personalized PageRank (PPR) \cite{page1999pagerank} or heat kernel \cite{kondor2002diffusion} along with graph sparsification to create the diffused version of the graph. One of the benefits of GDC is that it can perform multi-hop message-passing without requiring any specific change in the model architecture. 
Jin \textit{et al.} proposed Pro-GNN \cite{jin2020graph} that updates the graph structure iteratively with restrictions on the low-rank, sparsity, and feature smoothness properties.

Chen \textit{et al.} proposed AdaEdge \cite{chen2020measuring} that makes use of the homophily property of the graph and adds edges between nodes predicted to have the same label in the previous iteration and removes edges from the nodes having different labels. 
Similarly, Zhao \textit{et al.} proposed GAug-M and GAug-O \cite{zhao2021data} which adds/drops edges based on the output of neural edge predictor. GAug-M modifies the original graph for future training and inference whereas GAug-O modifies the graph for training only.  

Similarly, Multi-View Graph Convolution Network (MV-GCN) \cite{yuan2021semi} learns two new views of the graph data that reveal the global and feature relationship between nodes. Then an attention-based model aggregates the information from the three views (i.e. local topology, global topology, feature similarity) to learn node representation.

\subsubsection{Node Manipulation}
Node manipulation based methods goal is to create or remove nodes from the graph data. Inspired from the Mixup \cite{zhang2017mixup} data augmentation technique for images that merges two images to create a new image. 
Verma \textit{et al.} proposed GraphMix \cite{verma2019graphmix} as a replacement of Mixup for graph data. GraphMix is more like a regularization method that trains a fully-connected neural network along with GNN via parameter sharing. 
Following it Wang \textit{et al.} \cite{wang2021mixup} proposed Graph Mixup for node and graph classification. Graph Mixup is a two-branch convolution network. Given a pair of nodes, the two branches learn the node representation of each node and then the learned representations are aggregated at each hidden layer.

\begin{table*}[h]
    \centering
    \caption{Comparison of Data Augmentation Methods for Link Prediction with SOTAs. For GCN original architecture results in \textit{italics} are \% F1 Score.}

    \begin{tabular}{l|p{2cm}ccccccccc}
        \hline
        \textbf{GNN Arch.} & \textbf{Algorithms} & \textbf{Measure} & \textbf{Cora} & \textbf{Citeseer} & \textbf{Pubmed} & \textbf{Reddit} & \textbf{MOOC} & \textbf{Bitcoin Alpha} & \textbf{Bitcoin OTC}\\
        \hline
        
        \multirow{3}{4em}{GCN} & Original & AUC & 90.25 & 71.47 & 96.33 & - & - & \textit{64.00} &\textit{65.72}\\
        \cline{2-10}
        & +CFLP\cite{zhao2021counterfactual} & AUC & 92.55  & 89.65 & 96.99 & - &- &- &- \\
        & +GraphMix\cite{verma2019graphmix} & \% F1 Score & - & - &- & - & - & 65.34  & 66.35\\
        \hline
        \multirow{2}{4em}{TGN} & Original \cite{rossi2020temporal} & Accuracy &- & - & - & 92.56 &81.38& - & - \\
        \cline{2-10}
        & +MeTA\cite{wang2021adaptive} &  Accuracy &- & - & - & 94.19  &83.84 & - & -  \\
        \hline
        
    \end{tabular}
    \label{tab:edge-SOTAs}
\end{table*}

\subsubsection{Feature Manipulation}
Feature manipulation based techniques aim to manipulate the existing node features by either adding noise to them or by mixing up features of two or more nodes.

Kong \textit{et al.} proposed FLAG (Free Large-scale Adversarial Augmentation on Graphs) \cite{kong2020flag} which is not task-dependent and performs equally well on all the three tasks level (node, edge, and graph). The proposed algorithm iteratively augments node features by gradient-based perturbations. By making the model invariant to small perturbations the proposed model generalizes to out-of-distribution samples and hence gives better performance at test time. 
Local Augmentation for Graph Neural Networks (LA-GNN) \cite{liu2021local} learns neighborhood information by conditioning on local structures and node features. The learned features are then concatenated with the original raw features before feeding it to the GNN. 
Tang \textit{et al.} \cite{tang2021data} creates multiple new graph topology (adjacency matrices) and node attributes (feature matrices) and then feeds them to specific GCN models to get node embeddings. Final node embeddings are an attentional combination of all the node embeddings.

\subsubsection{Hybrid Manipulation}
Hybrid manipulation based methods are the combination of two or more augmentation techniques i.e. they perform augmentation by manipulating two or more graph components. 
Wang \textit{et al.} \cite{wang2020nodeaug} said that structural modification influences neighboring nodes and sometimes can lead to undesired results. To cater to this issue they introduced the concept of "\textit{Parallel Universe}". Structural modification of augmenting each node is made in a parallel universe in a way that the changes made for one node do not influence other nodes.
Feng \textit{et al.} \cite{feng2020graph} proposed GRAND for graph data augmentation. GRAND uses consistency loss to minimize the distance between the representation learning of different augmented graphs. The augmented graphs are generated by randomly masking out some node features and dropping a percentage of nodes.
Xue \textit{et al.} \cite{xue2021node} performs data augmentation by inserting new nodes into the graph. The labels and features of the inserted node are obtained by mixing the neighboring information with ratios $\lambda$ and $1-\lambda$. 

Sun \textit{et al.} proposed AutoGRL (Automated graph representation learning) \cite{sun2021automated} for node-level classification task. The proposed algorithm learns the best combinations of GDA operations along with GNN architecture and hyperparameters for the given dataset The proposed algorithm first designs appropriate search space and then finds the best combination of GDA operations, GNN architecture, and hyperparameters for the given dataset using efficient searching parameters. AutoGRL only has four GDA operations: Drop Node by randomly setting some node features to all zero, DropFeature by randomly setting some dimensions of node features to zero, Add/Drop Edges using GAE (Graph Auto Encoder) as in \cite{zhao2021data}, Label Nodes using LPA (Label Propagation Algorithm) \cite{zhou2003learning}.

\section{Graph Data Augmentation for Graph Classification}\label{sec:garph-aug}
\begin{table*}[ht]
    \centering
    \caption{Node-Level Tasks Data Augmentation Methods Accuracy(\%) Comparisons with SOTAs}

    \begin{tabular}{l|p{3cm} p{3cm} ccc}
        \hline
        \textbf{GNN Arch.} &\textbf{DA Technique} & \textbf{Algorithms} & \textbf{Cora} & \textbf{Citeseeer} & \textbf{Pubmed} \\
        \hline

        \multirow{11}{4em}{GCN}& \multirow{8}{1em}{Edge Augmentation}& Original\cite{kipf2016semi} & 81.50 & 70.30 & 79.00 \\
        \cline{2-6}
        & &+DropEdge\cite{rong2019dropedge}& 87.60 & 79.20 & 91.30 \\
        & &+NeuralSparse\cite{zheng2020robust} & 82.30 & 72.70 & - \\ 
        & &+TADropEdge\cite{gao2021training} & \textbf{88.30} & \textbf{81.40} & \textbf{91.50} \\
        & &+AdaEdge\cite{chen2020measuring} & 82.30 & 69.70 & 77.40 \\
        & &+GAug-M\cite{zhao2021data} & 83.50 & 72.30 & - \\ 
        & &+GAug-O\cite{zhao2021data}& 83.60 & 73.30 & - \\
        & & +MV-GCN\cite{yuan2021semi}& 83.60 & 72.90 & 81.50 \\
        \cline{2-6}
        & \multirow{2}{4em}{Node Augmentation}& +GraphMix \cite{verma2019graphmix}& 83.94 & 74.72 & 80.98 \\
        & &+Graph Mixup \cite{wang2021mixup}& \textbf{90.00} & \textbf{78.70} & \textbf{87.90}\\
        \cline{2-6}
        & \multirow{2}{4em}{Feature Augmentation} & +LA-GNN \cite{liu2021local} & 84.10 & 72.50 & 81.30 \\
        & & Tang \textit{et al.} \cite{tang2021data} & - & 74.60 & - \\
        \cline{2-6}
        & \multirow{3}{4em}{Hybrid Augmentation}& +NodeAug  \cite{wang2020nodeaug} & 84.30 & 74.90 & 81.50 \\
        & & +GRAND \cite{feng2020graph} & 84.50 & 74.20 & 80.00 \\
        & &+NodeAug-I  \cite{xue2021node}&82.10 &71.4& 78.8 \\
        \hline

        \multirow{8}{4em}{GAT} & \multirow{6}{4em}{Edge Augmentation} & Original\cite{velickovic2017graph} &83.00 & 72.50 & 79.00 \\
        \cline{2-6}
        & &+NeuralSparse\cite{zheng2020robust} & 84.20 & 73.60 & - \\ 
        & &+AdaEdge\cite{chen2020measuring} & 77.90 & 69.10 & 76.60 \\
        & &+GAug-M\cite{zhao2021data} & 82.10 & 71.50 & -  \\ 
        & &+GAug-O\cite{zhao2021data}& 82.20 & 71.60 & - \\
        \cline{2-6}
        & Node Augmentation & +GraphMix\cite{verma2019graphmix}& 83.32 & \textbf{73.08} & 81.10\\
        \cline{2-6}
        & Feature Augmentation & +LA-GNN\cite{liu2021local} & \textbf{83.90} & 72.30 & - \\
        \cline{2-6}
        & \multirow{2}{4em}{Hybrid Augmentation} &+NodeAug \cite{wang2020nodeaug}& \textbf{84.80} & \textbf{75.10} & \textbf{81.60}  \\
        & & +GRAND \cite{feng2020graph} & 84.30 & 73.20 & 79.20 \\
        & &+NodeAug\cite{xue2021node}&83.40 &70.8& 78.3 \\
        \hline

        \multirow{5}{4em}{GraphSAGE}& \multirow{5}{4em}{Edge Augmentation} & Original\cite{hamilton2017inductive} & 81.30 & 70.60 & 75.20 \\
        \cline{2-6}
        & &+DropEdge\cite{rong2019dropedge}& \textbf{88.10} & \textbf{80.00} & \textbf{91.70} \\
        & &+NeuralSparse\cite{zheng2020robust} & 84.10 & 73.60 & - \\
        & &+AdaEdge\cite{chen2020measuring} & 80.20 & 69.40 & 77.20\\
        & &+GAug-M\cite{zhao2021data} & 83.20 & 71.20& -  \\ 
        & &+GAug-O\cite{zhao2021data}& 82.00 & 72.70 & - \\
        \cline{2-6}
        & Hybrid Augmentation& +NodeAug\cite{xue2021node}& 82.30 &70.2& 78.1 \\
        \hline
    \end{tabular}
    \label{tab:node-SOTAs}
\end{table*}
Given a set of graphs denoted by $G=\{G^p\}_{p=1}^ n$, where $|G|=N$ and the associated label of $G^p$ is $y^p \in Y = \{1,2,...,K\}$, where K is the number of categories.  The goal of graph-level tasks is to predict the graph label of the unseen graph. A graph classifier C parameterized by $\theta$ can be written as $C_\theta (G) = y^o$, Here $y^o$ is the predicted graph label. The classifier is trained by following the objection function described in equation \ref{eq:graph_obj}, Where L(.,.) is the loss function to find a difference between true and predicted label and is usually cross-entropy loss.
\begin{equation}\label{eq:graph_obj}
  min_\theta \;\Sigma_i\; L(C_\theta(G_i), \;y_i)
\end{equation}

\subsection{Data Augmentation Methods}
Existing data augmentation methods for graph classification can be further divided into five classes: 1) Edge Manipulation 2) Node Manipulation 3) Feature Manipulation 4) Sub-Graph Manipulation 5) Hybrid Manipulation.
\subsubsection{Edge Manipulation}
Zhou \textit{et al.} \cite{zhou2020data} proposed M-Evolve algorithm that first finds motifs in the graph data and then adds/removes edges within the motifs based on sampling weights. 

\subsubsection{Node Manipulation}
Previously discussed Graph Mixup proposed by Wang \textit{et al.} \cite{wang2021mixup} a two-branch convolution network performs well for graph-level tasks as well. 

\subsubsection{Feature Manipulation}
Aforementioned, FLAG (Free Large-scale Adversarial Augmentation on Graphs) \cite{kong2020flag} which augments node features by iteratively adding gradient-based perturbations can also be used for graph-level tasks.

\subsubsection{Sub-graph Manipulation}
Sub-graph manipulation based techniques are for graph-level tasks only and in these techniques we extract a sub-graph from a graph and perform cropping, swapping and modification operation on that sub-graph. 
For graph-level tasks, since we have independent graphs it is comparatively easy to transfer some of the CV and NLP-based augmentation techniques to graphs. 
Inspired from image cropping augmentation technique, Wang \textit{et al.} proposed GraphCrop \cite{wang2020graphcrop}. It first randomly selects an initial node and then forms a sub-graph by retaining the nodes that hold the strongest connectivity with the initial node.  
Sun \textit{et al.} proposed MoCL \cite{sun2021mocl} designed specifically for biomedical domain. MoCL selects functional group from a graph and replaces them with other semantically similar functional group.

Guo and Mao proposed ifMixup \cite{guo2021ifmixup}, that directly apply mixup on graph data by randomly assigning indices to nodes and then matching nodes on the basis of those indices. 
Park \textit{et al.} proposed Graph Transplant \cite{park2021graph} which first selects a sub-graph from both source graph based on salient node and then transplant the source sub-graph into the destination graph. 
Similarly, Han \textit{et al.} proposed G-Mixup \cite{han2022g} for graph-level data augmentation. G-Mixup uses graphons to generate a class-level graph and then generate new data by mixing their graphons.
\begin{table*}[h]
    \centering
    \caption{Graph-Level Tasks Data Augmentation Methods Accuracy(\%) Comparisons with SOTAs}

    \begin{tabular}{l|p{3cm}p{3cm}ccccccc}
        \hline
        \textbf{GNN Arch.} &\textbf{DA Technique} & \textbf{Algorithms} & \textbf{Collab} & \textbf{IMDB-M} & \textbf{Reddit-5K} &\textbf{MUTAG} &\textbf{Proteins} & \textbf{NCI1}\\
        \hline
        \multirow{4}{4em}{GCN} &  & Original\cite{kipf2016semi} & 74.30 & 48.20 &\textbf{53.70}& 85.00 & 74.20 & 80.40\\
        \cline{2-9}
        & Node Augmentation & Graph Mixup\cite{wang2021mixup}& 75.40 & 48.80 & 54.60 & - &74.1 & 77.7\\
        \cline{2-9}
        & \multirow{6}{4em}{Subgraph Augmentation} & GraphCrop\cite{wang2020graphcrop}& 75.10 & 48.80 & 54.60 & - &74.0 & 77.3 \\
        & &G-Mixup \cite{han2022g}&- & \textbf{51.30}& 51.51 & - &- &-\\
        & &GAMS \cite{bicciato2022gams} &\textbf{75.47} & 47.80 & - & 90.70 & 73.78 & 65.82\\
        & &ifMixup \cite{guo2021ifmixup}& - & 52.30 & - & 87.90 & 75.30 & 81.90\\
        & & GraphCL \cite{you2020graph} & 74.23 & - & 52.55 & - & 74.17 & 74.63 \\
        & & JOAO \cite{you2021graph} & 75.53 & - & 52.83 & -& 73.31 & 74.86\\
        
        \hline
        \multirow{6}{4em}{GIN} & &Original \cite{xu2018powerful}& 75.50 & 48.50 &56.10 & 81.50 & 74.50 & 79.30\\
        \cline{2-9}
        & Node Augmentation &Graph Mixup\cite{wang2021mixup}& 77.00 & 49.90 & \textbf{57.8} & - & 74.3&81.0\\
        \cline{2-9}
        &\multirow{4}{4em}{Subgraph Augmentation} & GraphCrop\cite{wang2020graphcrop}& 76.9 & 49.60 & 57.7 & - & 74.1 & 80.9  \\
        & &G-Mixup \cite{han2022g}& - & \textbf{50.46} &  55.49 & - &- &-\\
        & &Graph Transplant \cite{park2021graph} &\textbf{81.50} & - & -& 82.80 & - & 81.50 \\
        & &ifMixup \cite{guo2021ifmixup}& - & 76.5 & - & 89.00 & 75.40 & 83.90 \\
        \hline

    \end{tabular}
    \label{tab:graph-SOTAs}
\end{table*}
 
Graph augmentation with module swapping (GAMS) \cite{bicciato2022gams} partitions the graph into a fixed number of clusters, then computes the similarity score between the clusters of two graphs and swaps the clusters that are most similar to each other.

\subsubsection{Hybrid Manipulation}
GraphCL \cite{you2020graph} augmentation algorithm carry out a combination of two or more augmentation techniques among the following techniques node augmentation (randomly remove nodes with their edges), edge augmentation (randomly add and drop edges), feature augmentation (mask out certain features randomly), and sub-graph augmentation(randomly sample a set of connected sub-graphs). Unfortunately, which combination of techniques to apply on a given data has to be hand picked by hit and errors because of  the diverse nature of graph data. 
To solve this limitation You \textit{et al.} proposed JOAO (JOint Augmentation Optimization) \cite{you2021graph} that automates the selection GDA techniques for GraphCL.

\section{Graph Data Augmentation for Link Prediction}\label{sec:edge-aug}
Given a dynamic graph $G =(V, E)$ in which each edge $e = (v_{i}, v_{j}) \in E$ represents an interaction between $v_{i}$ and $v_{j}$ that took place at a particular time $t$. We record multiple interactions between $v_{i}$ and $v_{j}$ as parallel edges, with potentially different timestamps. For two times $t < t^{'}$, let $G[t, t^{'}]$ denote the sub-graph of G consisting of all edges with a timestamp between $t$ and $t^{'}$. If we give our model $G[t_{0}, t^{'}_0]$; it must then output a list of edges not present in $G[t_{0}, t^{'}_0]$ that are predicted to appear in the network $G[t_{1}, t^{'}_1]$. 

\subsection{Data Augmentation Methods}
Very few dedicated GDA methods have been proposed for the task of link prediction. Existing works can be classified into three categories: 1)Edge augmentation 2)Node augmentation 3)Feature augmentation. 
\subsubsection{Edge Manipulation}
Wang \textit{et al.} proposed MeTA \cite{wang2021adaptive} for temporal graph. A multi-level module in MeTA handles augmented graphs of various magnitudes on multiple levels. The augmented graphs are created by perturbing time and edges. 
Zhao \textit{et al.} \cite{zhao2021counterfactual} method asks counterfactual questions of “would the link still exist if the graph structure became different from observation?” and learn to approximate unobserved outcome in the question. 

\subsubsection{Node Manipulation}
Aforementioned GraphMix proposed by Verma \textit{et al.} \cite{verma2019graphmix} works quite effectively for edge-level tasks as well. This method is inspired by Mixup \cite{zhang2017mixup} method from CV.
\subsubsection{Feature Manipulation}
Previously discussed FLAG \cite{kong2020flag} an adversarial attack based augmentation method works well for edge-level tasks. 

\section{Datasets and Evaluation Metric}\label{sec:data and eval}
In this section,  we will discuss the commonly used datasets  as well as evaluation metrics for assessment of data augmentation algorithms. 

\subsection{Datasets}
Table \ref{tab:node/edge data}, \ref{tab:graph-level data} summarizes statistics of datasets being used for data augmentation works on graph data.

\subsubsection{For Link Prediction}
To the best of our knowledge, till now there are only four works in literature that explicitly proposed data augmentation algorithms for edge-level tasks. Among the four works only two share a common dataset OGB-DDI. CFLP \cite{zhao2021counterfactual} used Cora, Citeseer, Pubmed \cite{sen:aimag08} citation network to evaluate their model performance. Where nodes are papers and edges represent citation links. 
Verma \textit{et al.} \cite{verma2019graphmix} used Bitcoin Alpha and Bitcoin OTC \cite{kumar2016edge} networks. The nodes in these networks represent bitcoin user and edge weight represent the level of trust between them. 
Wang \textit{et al.} \cite{wang2021adaptive} uses three temporal datasets: MOOC \cite{kumar2019predicting}, Reddit \cite{baumgartner2020pushshift}, and Wikipedia \cite{kumar2019predicting}. 
Reddit is a social network where a node is a user and an edge represents their interaction on the reddit thread. 
MOOC is a temporal and directed user action network. In MOOC networks, nodes are users and course activities whereas edge represents action by users on course activities. 
Similarly, Wikipedia is a temporal network of Wikipedia page edits. An edge represents that the user edited the page at a specific time.

\subsubsection{For Node Classification}
Cora, Citeseer, and Pubmed \cite{sen:aimag08} are the three commonly used datasets to evaluate data augmentation algorithms for node-level tasks. 
All three datasets are undirected citation networks where nodes are documents and edges are citation links. The features are sparse bag-of-words vectors. For training, only 20 nodes per class were used i.e. 140 nodes for Cora, 120 nodes for Citeseer, and 60 nodes for Pubmed.

\subsubsection{For Graph Classification}
For Graph-level task Collab \cite{yanardag2015deep}, IMDM-M, Reddit-5K \cite{baumgartner2020pushshift},  Proteins \cite{KKMMN2016}, MUTAG \cite{kazius2005derivation},and NCI1 \cite{wale2008comparison} undirected networks are being widely used for evaluating model performance.  
Collab is a scientific collaboration network where nodes in each graph are researchers and the edge indicates the collaboration between two researchers. 
IMDM-M is a co-featuring network of actors/actresses where nodes in each graph are actors/actresses and the edge indicates that the actor/actress worked in the same movie. '
Reddit-5K is a social network where each graph is a discussion thread. Nodes in each graph are users and an edge is showing whether a user responded to another user or not. 
Proteins is datasets of proteins and the goal is to classify proteins as enzymes and non-enzymes. Node represents the amino acids and edge represents that the nodes/amino acids are less than 6 Angstroms apart. 
MUTAG is a collection of nitroaromatic compounds where each graph represents a chemical compound. The nodes in the graph are atoms and edge represents bond between atoms. 
Similarly, NCI1 is also a collection of chemical compounds where a graph is a chemical compound. Nodes represents atoms and edges denote the bond between the atoms. NCI1 dataset is relative to anti-cancer screens where the chemicals are assessed as positive or negative to cell lung cancer.

\subsection{Evaluation Metrics}
In this section, we summarize the commonly used evaluation measures in the data augmentation for graph data literature. For convenience, we divide them on the the basis of end task.
\subsubsection{For Link Prediction}
In this section, we briefly describe commonly used evaluation metrics for edge-level tasks. We could not find a common evaluation metric among the edge-level tasks prior works and hence will report the evaluation metric used in each paper separately. Verma \textit{et al.} \cite{verma2019graphmix} uses F1-score as evaluation metric. Zhao \textit{et al.} used AUC as the link prediction evaluation measure in their work\cite{zhao2021counterfactual}. Wang \textit{et al.} in their work \cite{wang2021adaptive} , evaluated their model performance by test accuracy.
\subsubsection{For Node Classification}
Most of the existing data augmentation algorithms  \cite{rong2019dropedge,chen2020measuring, kong2020flag,wang2020nodeaug,tang2021data,klicpera2019diffusion} for node-level tasks use classification accuracy as an evaluation metric. Apart from accuracy,  \textbf{F1-Score} \cite{wiki:f1} and Area-Under-the-ROC-Curve \textbf{AUC} \cite{wiki:roc}  is also used by some works such as \cite{wang2021mixup,  zheng2020robust}. Value of accuracy for an algorithm must lie between 0 to 100 and values close to 100 represent superior performance whereas value of F1-score and  AUC must lie in the range 0 to 1 and a value of 1 represents 100\% perfect prediction. 

\subsubsection{For Graph Classification}
Existing data augmentation algorithms \cite{wang2020graphcrop, you2020graph, you2021graph, sun2021mocl, guo2021ifmixup, bicciato2022gams} for graph-level tasks mostly use classification accuracy as performance evaluation metric except for few. MoCL \cite{sun2021mocl} that is using Area-Under-the-ROC-Curve \textbf{AUC} \cite{wiki:roc} as evaluation metric. Similarly, FLAG \cite{kong2020flag} uses Area-Under-the-ROC-Curve \textbf{AUC}\cite{wiki:roc}, F1-score \cite{wiki:f1} and Average precision

\section{Performance Comparison}
In this section we compare the performance of existing graph data augmentation methods on the commonly used datasets for each task-level separately. For fair comparison the results are directly taken from the respective original paper.
\label{sec:perform comp}
\subsection{For Link Prediction}
In this section, we compare the performance of the proposed DA techniques against state-of-the-arts (SOTAs) GNN architectures  GCN \cite{kipf2016semi}, and TGN \cite{rossi2020temporal}. Table \ref{tab:edge-SOTAs}, compare the performance of edge-level DA techniques with SOTAs. CFLP \cite{zhao2021counterfactual} DA technique improves AUC of GCN \cite{kipf2016semi} by 2.3\%, 18.18\%, and 0.66\% on Cora, Citseer, and Pubmed datasets respectively. GraphMix \cite{verma2019graphmix} DA technqiue shows slight improvement of 0.98\% (averaged over both datasets) in \% F1 score for Bitcoin Alpha and Bitcoin OTC datasets in comparison to original GCN \cite{kipf2016semi}. MeTA \cite{wang2021adaptive} showed an improvement of 1.63\% in test accuracy as compared to TGN \cite{rossi2020temporal} on Reddit dataset.

\subsection{For Node Classification}
This section compares the performance of the existing data augmentation methods for node-level tasks against state-of-the-arts (SOTAs) GNN architectures  GCN \cite{kipf2016semi}, GAT \cite{velickovic2017graph}, and GraphSAGE \cite{hamilton2017inductive}. Table \ref{tab:node-SOTAs}, provides the accuracy (\%) comparison of existing data augmentation methods with SOTAs. 

\textbf{Edge Manipulation based DA Methods:} 
For the GCN \cite{kipf2016semi} architecture, all the proposed DA methods showed an improvement in accuracy on all the three datasets except for the Pubmed dataset in the case of AdaEdge \cite{chen2020measuring} for which we observed a 1.6\% decrease instead. 
TADropEdge \cite{gao2021training} proposed by Gao \textit{et al.} gave the best performance on all three datasets for GCN architecture. For Cora, Citeseer, and Pubmed an increase of 6.8\%, 11.1\%, and 12.5\% respectively was observed. 
For the GAT \cite{velickovic2017graph} GNN architecture, NeuralSparse  \cite{zheng2020robust} gave the best performance. Similarly, for GraphSAGE \cite{hamilton2017inductive} GNN architecture DropEdge \cite{rong2019dropedge} gave the best performance. For GCN, we observed an overall 10.13\%  increase in the accuracy(averaged over all the datasets). For GAT, an overall 5.27\% increase in accuracy was recorded (averaged over all the datasets). Similarly, for GraphSAGE there was an overall 10.9\% increase in accuracy (again averaged over all the datasets).

\textbf{Node Manipulation based DA Methods:}
For the GCN \cite{kipf2016semi} architecture, Graph Mixup \cite{wang2021mixup} augmentation technique showed an average 8.6\% increase in accuracy across the three datasets.  whereas as GraphMix \cite{verma2019graphmix} performed well on Citeseer dataset and showed a 0.58\% increase in accuracy. For GraphSAGE \cite{hamilton2017inductive}, DropEdge \cite{rong2019dropedge} gave best performance on all the three datasets and showed an average increase of 10.9\% in test accuracy.

\textbf{Feature Manipulation based DA Methods:} 
For GCN \cite{velickovic2017graph}, Tang \textit{et al.} data augmentation method showed an increase of 4.3\% on citeseer dataset and an increase of 2.6\% was observed on cora dataset for LA-GNN  \cite{liu2021local} data augmentation method.
For GAT \cite{velickovic2017graph}, LA-GNN \cite{liu2021local} was able to increase testing accuracy for Cora dataset by 0.9\%.

\textbf{Hybrid Manipulation based DA Methods:}
For the GCN \cite{kipf2016semi} architecture, NodeAug \cite{xue2021node} proposed methods showed slight improvement in accuracy for Cora and Citeseer dataset. However, for Pubmed, the accuracy decreased by 0.2\%. For GAT \cite{velickovic2017graph}, NodeAug accuracy improved by 0.4\% for only Cora dataset. For both Citeseer and Pubmed the accuracy declined. Whereas in the case of GraphSAGE \cite{hamilton2017inductive} GNN architecture, NodeAug couldn't perform better than the original GraphSAGE.

\subsection{For Graph Classification}
In this section, we compare the performance of the proposed DA techniques against state-of-the-art (SOTAs) GNN architectures  GCN \cite{kipf2016semi}, and GIN \cite{xu2018powerful}. Table \ref{tab:graph-SOTAs}, provides the accuracy (\%) comparison of proposed DA techniques with SOTAs. For graph-level tasks, a single augmentation technique failed to perform well on all the datasets. In the case of GCN \cite{kipf2016semi} GNN architecture, GAMS \cite{bicciato2022gams} performed best on Collab dataset, G-Mixup \cite{han2022g} gave the best results for IMDB-M dataset whereas all the discussed data augmentation techniques were not able to show any improvement in case of Reddit-5K dataset. For GIN \cite{xu2018powerful} architecture, Graph Transplant \cite{park2021graph} showed a 6\% increase in performance for Collab dataset, G-Mixup \cite{han2022g} performed well on IMDB-M dataset with an improvement of 1.96\% in test accuracy, and Graph Mixup \cite{wang2021mixup} showed slight performance improvement for Reddit-5k dataset.

\section{Conclusions and Future Directions}
\begin{table*}[h]
    \centering
    \caption{Summary of open-source implementation of algorithms}

    \begin{tabular}{l|p{1cm}p{3cm}p{6cm}}
        \hline
        \textbf{Task Level}& \textbf{Paper} & \textbf{Algorithm} & \textbf{Link} \\
         \hline
         \multirow{10}{6em}{\textbf{Node-level}} & \cite{rong2019dropedge} & DropEdge & \url{https://github.com/DropEdge/DropEdge}\\
         
         & \cite{zhao2021data} & GAug-M, GAug-O  & \url{https://github.com/zhao-tong/GAug}\\
         
         & \cite{zheng2020robust} & NeuralSparse & \url{https://github.com/flyingdoog/PTDNet}\\
         
         & \cite{jin2020graph} & Pro-GNN & \url{https://github.com/ChandlerBang/Pro-GNN}\\
         
         & \cite{kong2020flag} & FLAG & \url{https://github.com/devnkong/FLAG}\\
         
         & \cite{liu2021local} & LA-GNN & \url{https://github.com/Soughing0823/LAGNN}\\
         
         & \cite{klicpera2019diffusion} & GDC & \url{https://github.com/gasteigerjo/gdc}\\
         
         & \cite{verma2019graphmix} & GraphMix & \url{https://github.com/vikasverma1077/GraphMix}\\
         
        & \cite{wang2021mixup} & Graph Mixup & \url{https://github.com/vanoracai/MixupForGraph}\\
         
         & \cite{sun2021automated} & AutoGRL& \url{https://github.com/JunweiSUN/AutoGRL}\\
        
         & \cite{feng2020graph} & GRAND & \url{https://github.com/Grand20/grand}\\
         & \cite{velicokovic2018deep} & DGI & \url{https://github.com/PetarV-/DGI}\\
         & \cite{zhu2020deep} & GRACE & \url{ https://github.com/CRIPAC-DIG/GRACE}\\
          \hline
         \multirow{4}{6em}{\textbf{Graph-level}} & \cite{sun2021mocl} & MoCL & \url{https://github.com/illidanlab/MoCL-DK}\\
        & \cite{bicciato2022gams} & GAMS & \url{https://gitlab.com/ripper346-phd/graph-cluster-}\\
        &\cite{you2020graph} & GraphCL &\url{https://github.com/Shen-Lab/GraphCL}\\
        & \cite{you2021graph} & JOAO &  \url{https://github.com/ Shen-Lab/GraphCL_Automated}\\
        \hline

        \textbf{Edge-level}& \cite{zhao2021counterfactual} & CFLP & \url{https://github.com/DM2-ND/CFLP}\\
        \hline
        
    \end{tabular}
    \label{tab:links}
\end{table*}
In this work, we compare all the existing graph data augmentation techniques to the best of our knowledge. We divide the existing works on the basis of task-levels and briefly describes the existing methods for each task-level. Then, we share most commonly used datasets and evaluation metrics for each task-level. Later, we also provide performance comparison of existing GDA technqiues on commonly used datasets. 

\subsection{Future Directions}
We now summarize a few limitations we have found in this field of research and offer some suggestions for the future works:
\begin{itemize}
    \item Most of the existing data augmentation methods focus on node-level (especially node classification) and graph-level (especially graph classification) tasks and very few attention is paid to edge-level tasks. There is also opportunity in real-life tasks such as graph-based recommendation, search etc.
    
    \item To the best of our knowledge, FLAG proposed by Kong \textit{et al.} \cite{kong2020flag} is the only GDA algorithm that is not specified to a particular task-level i.e node-level, edge-level, and graph-level and performs equally well on all the three task-levels. There is a potential for more generic algorithms like FLAG.

    \item Most of the existing data augmentation works cannot be used for molecular data because randomly adding/removing edges or cropping a subset of graph may lead to invalid molecular structure. To mitigate this issue their is a need to come up with more generalized graph data augmentation methods.
    
    \item Most existing works do not study the efficiency and scalability of graph data augmentation methods on larger networks. As we know that real-world network can be massive so this aspect cannot be ignored.
    
\end{itemize}

\section{Appendix}
\label{appendix}
In this work, we not only survey the existing GDA techniques but also provide the details of commonly used datasets and evaluation metrics. Moreover, in table \ref{tab:links} we provide a link to open-source implementation in hope of facilitating the research community. 
\bibliography{gda.bib} 
\bibliographystyle{plain}

\end{document}